\documentclass{article}

\usepackage[preprint]{neurips_2024}
\usepackage{listings}
\usepackage{xfrac}    
\usepackage[most]{tcolorbox}
\usepackage{amssymb}
\usepackage{color}
\usepackage{multirow}
\usepackage{float}
\definecolor{keywordcolor}{rgb}{0.7, 0.1, 0.1}   
\definecolor{tacticcolor}{rgb}{0.0, 0.1, 0.6}    
\definecolor{commentcolor}{rgb}{0.4, 0.4, 0.4}   
\definecolor{symbolcolor}{rgb}{0.0, 0.1, 0.6}    
\definecolor{sortcolor}{rgb}{0.1, 0.5, 0.1}      
\definecolor{attributecolor}{rgb}{0.7, 0.1, 0.1} 

\usepackage[utf8]{inputenc} 
\usepackage[T1]{fontenc}    
\usepackage{hyperref}       
\usepackage{url}            
\usepackage{booktabs}       
\usepackage{amsfonts}       
\usepackage{nicefrac}       
\usepackage{microtype}      
\usepackage{xcolor}         
\usepackage{pifont}
\usepackage{wrapfig}
\newcommand{\dataset} {LEAN-GitHub}
\newcommand{\prover} {InternLM2-StepProver }
\title{\dataset{}: Compiling GitHub LEAN repositories for a versatile LEAN prover}

\author{%
    Zijian Wu$^{12}$\thanks{Work done during internships at Shanghai AI Laboratory.}, Jiayu Wang$^{1}$, Dahua Lin$^{1}$, Kai Chen$^{1}$ \\
    $^{1}$Shanghai AI Laboratory, $^{2}$The Chinese University of Hong Kong\\
    \texttt{wuzijian@pjlab.org.cn}
}

\begin{document}

\maketitle

\begin{abstract}
  %
Recently, large language models have presented promising results in aiding formal mathematical reasoning. However, their performance is restricted
due to the scarcity of formal theorem-proving data, which requires additional effort to be extracted from raw formal language corpora. Meanwhile,
a significant amount of human-written formal language corpora remains underutilized. To address this issue, we propose \dataset{}, a dataset consisting of large-scale
formal data extracted from almost all Lean 4 repositories on GitHub. After fine-tuning InternLM-math-plus on this dataset, our model
achieved accuracies of 48.8\% with a single pass and 54.5\% with 64 passes on the Lean 4 miniF2F test, surpassing state-of-the-art method at
52\%. And it also achieves state-of-the-art on two other Lean 4 benchmarks (ProofNet and Putnam) targeting different fields/levels of math. These results demonstrate that our proposed dataset is beneficial
for formal reasoning on a wide range of math topics. 
We open-source our model at \url{https://GitHub.com/InternLM/InternLM-Math} and our data at \url{https://huggingface.co/datasets/InternLM/Lean-GitHub}.
\end{abstract}
\def\lstlanguagefiles{lstlean.tex}
\section{Introduction}
Theorem proving stands as a fundamental objective in mathematics. To tackle the escalating intricacy of proofs and identify non-trivial flaws within them, formalized mathematical systems like Lean~\cite{de2015lean}, Isabelle~\cite{paulson_isabelle_1994}, and Coq~\cite{coq} have been developed to furnish computer-verifiable proofs~\cite{avigad2023mathematics}. However, crafting formal proofs demands substantial human effort, posing challenges for further advancement and underscoring the necessity for automated theorem proving~\cite{machinelogic1956}. Recently, large language models (LLMs)~\cite{achiam2023gpt,azerbayev2023llemma,yang2024leandojo,polu2020generative,thakur2023language,han2021proof,wang2024theoremllamatransforminggeneralpurposellms,ying2024internlmmathopenmathlarge} have shown promising results in resolving high-school level math problems through interactions with formalized proof assistants. Nevertheless, their performance remains unsatisfactory, primarily due to data scarcity.

Formal languages necessitate significant expertise and effort and are utilized by a relatively small number of mathematicians, leading to a shortage of formal language corpora. In addition, unlike conventional programming languages such as Python or Java, formal proof languages contain intermediate information not directly visible in their raw code, e.g. proof trees comprising intermediate states between proof steps, making raw language corpora unsuitable for training. This scarcity of well-crafted human-written formal language data persists while many valuable human-written corpora remain underutilized. Although auto-formalization~\cite{xin2024deepseekproveradvancingtheoremproving,ying2024leanworkbooklargescalelean} enables the synthesis of more aligned data for training, the quality and diversity of their data remain constrained and thus cannot substitute for human-crafted data.

\begin{figure}[h]
    \centering
    \begin{minipage}[t]{0.68\textwidth}
        \begin{tcolorbox}[
            colback=white!10!white,
            colframe=purple!75!purple,
            title=Case: IMO 1983 P6
        ]
        \textcolor{blue}{Natural Language problem:} Let $a$, $b$ and $c$ be the lengths of the sides of a triangle. Prove that\[a^{2}b(a - b) + b^{2}c(b - c) + c^{2}a(c - a)\ge 0.\] 
        \\
        \begin{lstlisting}[language=LEAN]
theorem imo_1983_p6 (a b c : ℝ) (h₀ : 0 < a ∧ 0 < b ∧ 0 < c) (h₁ : c < a + b) (h₂ : b < a + c)
    (h₃ : a < b+c) : 0 ≤ a^2 * b * (a-b) + b ^ 2 * c * (b - c) + c ^ 2 * a * (c - a) := 
by
    ring_nf
    have h₄ : 0 < a+b+c := by linarith
    simp only [add_assoc]
    have h₅ : 0 ≤ (a-b)^2 * (a+b-c) := by nlinarith
    have h₆ : 0 ≤ (a-c)^2 * (a+c-b) := by nlinarith
    have h₇ : 0 ≤ (b-c)^2 * (b+c-a) := by nlinarith
    nlinarith [h₀.1, h₀.2.1, h₀.2.2, h₁, h₂, h₃, h₄, h₅, h₆, h₇]
        \end{lstlisting}
        \label{tab:imo_case_top}
        \end{tcolorbox}
        \caption{An IMO problem solved by \prover.}
    \end{minipage}
    \hfill
    \begin{minipage}[t]{0.28\textwidth}
        \begin{center}
            \includegraphics[width=\textwidth]{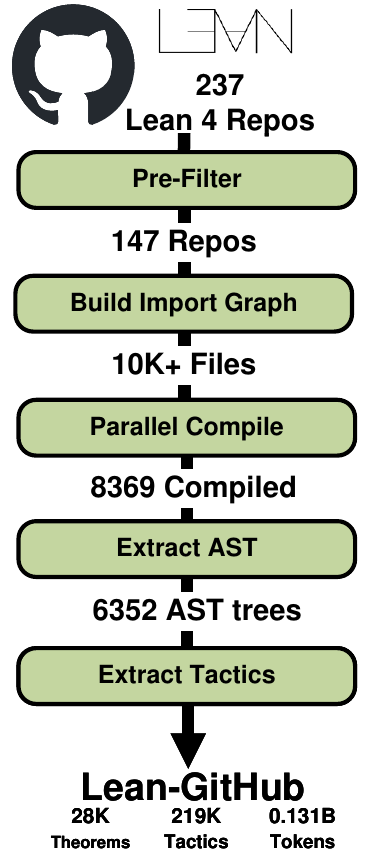}
        \end{center}
        \caption{The pipeline for constructing \dataset{}.}
        \label{fig:pipeline}
    \end{minipage}
\end{figure}

To address this challenge, we propose \dataset{} in this paper: a large-scale Lean dataset that leverages open-source Lean repositories on GitHub, serving as a crucial complement to the well-utilized Mathlib~\cite{mathlib,yang2024leandojo} dataset. We develop a scalable pipeline, shown in Fig.~\ref{fig:pipeline}, to boost the extraction efficiency and parallelness, and managed to exploit precious data from Lean corpus that were not compiled and extracted before. We also provide a solution to the state duplication problem common in tree proof search methods. To showcase the efficacy of our dataset, we train \prover with our dataset included. Quantitative results show that fine-tuning on our dataset enhances formal reasoning abilities in Lean 4 across various formal benchmarks, indicating our proposed dataset is beneficial for formal reasoning on versatile math topics.

In summary, our paper makes the following contributions:
\begin{itemize}

\item We publicly release \dataset{}, a dataset consisting of large-scale formal data extracted from open Lean 4 repositories on GitHub, including 28,597 theorems and 218,866 tactics, fostering further research and development in automated theorem proving. Our dataset could be found at \url{https://huggingface.co/datasets/InternLM/Lean-GitHub}.
 \item Our model \prover, a 7B model trained on this dataset, achieves state-of-the-art performance on
 benchmarks targeting different fields/levels of math. The \prover \ has reported accuracies of 48.8\% within a single pass and 54.5\% with 64 passes on the Lean 4 miniF2F~\cite{zheng2021minif2f} test, surpassing Deepseek-Prover~\cite{xin2024deepseekproveradvancingtheoremproving} at 52\%; an 18.1\% Pass@1 on ProofNet~\cite{azerbayev2023proofnetautoformalizingformallyproving}; and has solved 5 out of 640 Putnam~\cite{tsoukalas2024putnambenchevaluatingneuraltheoremprovers} problems. In addition, our model solved 3 IMO problems in the miniF2F dataset, see Fig.~\ref{tab:imo_case_top} for an example.\footnote{The original problem includes determining when equality occurs. It is omitted in our formalization due to the restriction of the formal system. }  
\end{itemize}


\section{Related works}

 The development of modern proof assistants, including Coq~\cite{coq}, Isabelle~\cite{paulson_isabelle_1994}, and Lean~\cite{de2015lean}, has expanded the expressive capabilities of formal systems beyond first-order logic, thereby stimulating heightened interest in automated theorem proving. The recent integration of large language models~\cite{achiam2023gpt,azerbayev2023llemma,shao2024deepseekmath,xin2024deepseekproveradvancingtheoremproving,ying2024internlmmathopenmathlarge} has further advanced the development of tools and datasets.

\textbf{Automatic Theorem Proving}
Earlier works in ATP use traditional methods, such as KNN~\cite{Gauthier2018TacticToeLT} or GNN~\cite{Yang2019LearningTP}. Some~\cite{kaliszyk2018reinforcement,crouse2021deep,wu2021tacticzero} exploit reinforcement learning techniques to improve performance. Recently, more are taking advantage of the deep transformer-based methods that
 treat theorems as plain texts. Many learning-based theorem proving, such as GPT-f~\cite{polu2020generative}, PACT~\cite{han2021proof}, Llemma~\cite{azerbayev2023llemma}, COPRA~\cite{thakur2023language},  ReProver~\cite{yang2024leandojo} and Lean-STaR~\cite{lin2024leanstarlearninginterleavethinking}, instructs a language model on (proof state, next-tactic) pairs, then proves theorems by tree search. Some~\cite{Yang2019LearningTP} tries generating tactics on the granularity of abstract syntax trees. An alternative approach involves harnessing LLMs to autonomously generate the entire proof either independently or based on human-provided proofs~\cite{xin2024deepseekproveradvancingtheoremproving,first2023baldur,zhao2023decomposing,jiang2022draft,xin2023lego}. Some other methods also explores the possibility of incorporating informal proofs into formal proofs~\cite{jiang2022draft,wu2022autoformalization,lin2024leanstarlearninginterleavethinking,wang2024theoremllamatransforminggeneralpurposellms}. We follow the GPT-f framework and trains our model on the granularity of tactics.

\textbf{Data Extraction for Lean Corpus}
Following human-behavior in interacting with proof assistants, it is crucial for
the automated theorem provers to see intermediate states invisible in the code but visible to human at runtime. Therefore, data extraction tools are critical drivers of ATP: Coq has GamePad~\cite{huang2018gamepadlearningenvironmenttheorem} and CoqGym~\cite{yang2019coqgymlearningprovetheoremsinteracting}; Isabelle has IsarStep~\cite{li2021isarstepbenchmarkhighlevelmathematical}, and Lean has LeanStep~\cite{han2021proof}, lean-gym~\cite{polu2022formal} (not compatible with Lean 4) and LeanDojo~\cite{yang2024leandojo}. We focus on extraction tools for Lean4. 
However, prior Lean 4 tools involve significant overhead in extraction as it is designed for a single project, therefore not directly suitable for massive extraction on multiple projects.


\section{The GitHub-Lean dataset}
There are numerous Lean repositories on GitHub, which contains scarce human-written theorems and proofs. However, the raw Lean code is unsuitable for direct training, as it has crucial runtime
 information that humans can access when interact with Lean environments unrevealed. Examples include the intermediate states and targets between each proof steps, and hints provided by some tactics. 
 Though there has been works on extracting these information, they are restricted on Mathlib 4~\cite{yang2024leandojo,mathlib}, Lean's centralized formal math library. Meanwhile, hundreds of Lean repositories covering diverse topics exists on GitHub without getting exploited and extracted.

We form a dataset for theorem proving in Lean 4, named \dataset{}, built upon 147 Lean 4 repositories available on the web.
The dataset is one of the largest theorem proving datasets in Lean 4 formal mathematics, consisting of 28,597 theorems with formal proofs and 218,866 tactics from 2133 files. The dataset has 0.138B tokens. We propose that, training on the dataset improves model performance on theorem proving in various mathematical topics.
\subsection{Dataset Construction}
In this section, we will detailedly describe how we construct the \dataset{} dataset. 
\textbf{Selection of the repositories.} After conducting an exhaustive search on GitHub, we identified a total of 237 Lean 4 repositories (GitHub does not differentiate between Lean 3 and Lean 4) which may contain compilable theorems. By filtering for keywords such as "theorem" and "lemma", we estimated that there are approximately 48,091 theorems across these repositories. However, it is important to note that the presence of a keyword does not guarantee the existence of a theorem with a proof written in the form of tactics. The main obstacles include: a) some repositories cannot compile, either due to improper construction of the project or incorrect Lean files included; b) dependencies on other repositories that are not available online; c) repositories written in older versions of Lean with deprecated features that cannot be migrated to newer versions; and d) proofs of theorems not constructed using tactics. We discarded 90 repositories written in deprecated Lean 4 versions.

Among the remaining repositories, only 61 out of 147 could be compiled as valid Lean 4 projects without requiring modifications. 
The remaining repositories required extra efforts to compile successfully. A small fraction of projects relies on non-official releases of Lean 4, others contains a significant number of isolated files. We develop automated scripts to try heuristically finding the closest official releases for the former case. Solution to the latter case is described in the next paragraph.

\textbf{Source Code Compilation.} We opted not to use the \texttt{Lake}~\cite{de2015lean} tool provided by the Lean 4 standard library, but instead called the underlying \texttt{leanc} compiler directly to compile the source code. This approach offers two advantages. First, many of the Lean repositories collected are not compliant Lean projects and cannot be compiled. This is because Lean 4 can function as both a compiled language and a script language. Mathematicians often tend to write isolated files within an empty Lean project, which cannot be compiled by \texttt{Lake}. Second, \texttt{Lake} would fail to build the project if any of its building targets failed, causing the content of the whole project to be discarded. We also observed a performance bottleneck in \texttt{Lake}'s concurrent primitives. To address this issue, we first extended \texttt{Lake}'s \textit{import graph} on file dependencies. We modified \texttt{Lake} to expose this information. Then we augmented it with information of isolated files and rebuild our global \textit{import graph}. With this dependency information, we could replace \texttt{Lake} with a custom compiling script that directly calls the underlying \texttt{leanc} compiler with increased parallelness.

\textbf{Extraction Details.} We develop our extraction utilities based on LeanDojo~\cite{yang2024leandojo}. Tools such as LeanDojo~\cite{yang2024leandojo} and LeanStep~\cite{han2021proof} typically require the entire project to be compiled before data extraction. We argue that this restriction is unnecessary and implemented data extraction for isolated files. Observing some bottlenecks introduced by LeanDojo's reliance on network connection and its design choice that putting data extraction and interaction with Lean together, which brings many computational redundancies, we restructured the implementation with an increased parallelism. Out of 8639 Lean source files, 6352 files and 42K theorems were successfully extracted, with 2133 files and 28K theorems containing valid tactic information. 
\subsection{Dataset Statistics}

\begin{figure}[t]
	\centering
	\begin{minipage}[t]{0.46\textwidth}
		\centering
        \includegraphics[width=1.3\linewidth]{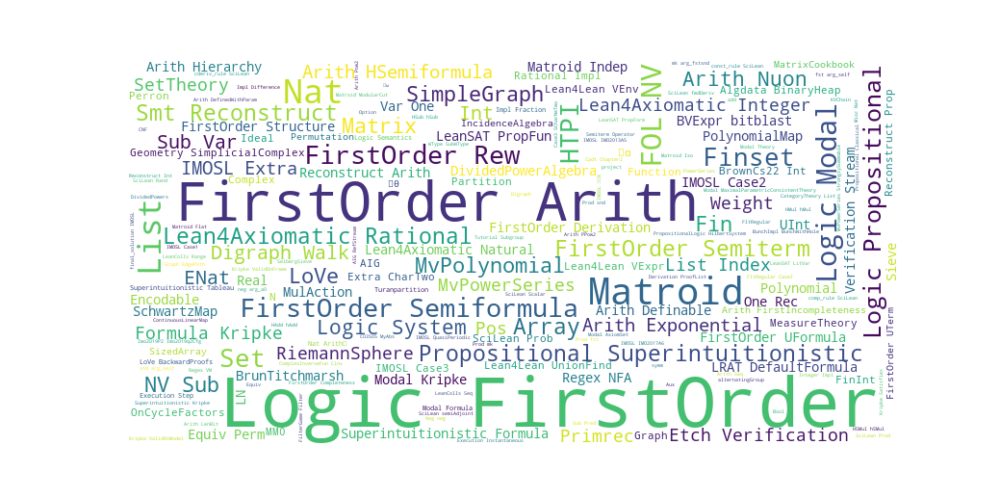}
		\caption{Word Cloud for the theorem names in \dataset{}.}
		\label{fig:word-cloud}
	\end{minipage}\hfill
	\begin{minipage}[t]{0.48\textwidth}
		\centering
		\includegraphics[width=0.95\linewidth]{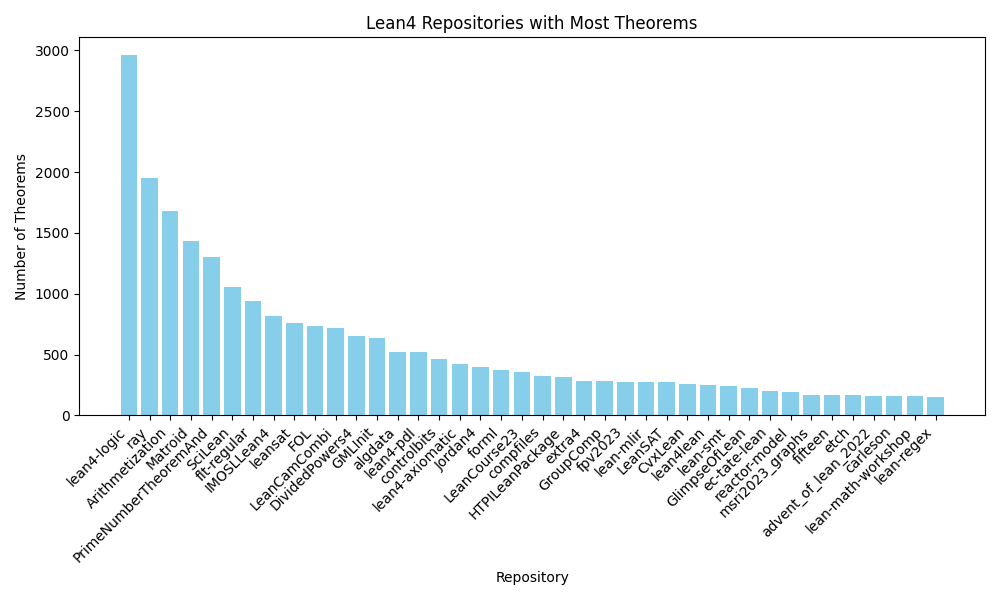}
		\caption{Top 30 repositories with most theorems extracted.}
		\label{fig:theorem-distribution}
	\end{minipage}
	\vspace{-0.15in}
\end{figure}

Fig.~\ref{fig:word-cloud} shows the word cloud of the set of theorem names of our dataset. The word cloud
 highlights the most frequently occurring keywords in the theorem names, with "Logic", "FirstOrder", "Matroid", and "Arith(mezation)" being the most prominent, indicating that the dataset contains mathematics from various fields. Fig.~\ref{fig:theorem-distribution} displays the top 30 repositories with the most theorems extracted. The distribution further shows the diverse mathematical topics in the data, including cutting-edge mathematical fields, data structures, as well as Olympiad-level problems.

 \begin{table}[hb]
	\centering
	\resizebox{0.99\textwidth}{!}{
		\begin{tabular}{l|ccccc}
			\toprule
			\textbf{Dataset} & \textbf{Lean-Workbook~\cite{ying2024leanworkbooklargescalelean}} & \textbf{Deepseek-Prover~\cite{xin2024deepseekproveradvancingtheoremproving}} & \textbf{miniF2F-curriculum}~\cite{polu2022formal} & \textbf{LeanDojo-Mathlib~\cite{yang2024leandojo}} & \textbf{\dataset{}  (ours)} \\
			\midrule
			\textbf{Open-sourced} & \checkmark & \ding{53} & \checkmark & \checkmark & \checkmark \\
			\textbf{Source} & Synthetic & Synthetic & Human-written & Human-written & Human-written \\
			\textbf{Intermediate States} & \ding{53} & \ding{53} & \ding{53} & \checkmark & \checkmark \\
			\textbf{Theorems} & 57K & 870K & 327 & 60K & 28K \\
			\textbf{Tokens} & 0.029B & 3.108B & 1.5K & 0.138B & 0.131B \\
			\textbf{Level} & Undergraduate & Undergraduate & High-school & Diverse & Diverse \\
			\bottomrule
		\end{tabular}
	}
	\caption{Comparison of dataset statistics among Lean 4 formal reasoning datasets: Lean-Workbook, Deepseek-Prover, miniF2F-curriculum, LeanDojo-Mathlib, and \dataset{}  (ours).}
	\label{tab:comparison}
\end{table}
To estimate the quality our dataset, we provide several quantitative measures and compare them with recent and similar existing datasets, namely Lean-Workbook, Deepseek-Prover, miniF2F-curriculum and LeanDojo-Mathlib (see Tab.~\ref{tab:comparison}). 

Lean-Workbook and Deepseek-Prover are synthetic datasets whose topics are largely restricted to high-school level (some in the undergraduate level), and crucial intermediate steps are not available. Besides, since they rely on present methods to generate solutions, their solution length and quality are also restricted. 
MiniF2F-curriculum, designed for expert iteration~\cite{polu2022formal} has a small size of 327 examples, limiting its versatility and robustness. 
LeanDojo-Mathlib is one of the largest datasets at the granularity of tactics, extracted from Mathlib 4, the standard mathematics library for Lean 4. It focuses mainly on describing specific mathematical theories rather than the topic of problem-solving.
In contrast, our dataset, which contains versatile fields, levels, and tastes of math proofs, is comparably large and covers a broad spectrum of complexities.
Overall, our dataset pushes the boundaries of utilizing precious human-written corpora and is comparable to prior efforts.


\section{Experiments}
We develop the \prover model that utilizes the \dataset{}. \prover is built upon InternLM-math-plus-7B~\cite{ying2024internlmmathopenmathlarge} model, which is a decoder-only transformer that is continued pre-trained on a corpus comprising 200B informal and formal math-related tokens. We then conduct extensive experiments on various Lean 4 datasets to test the effectiveness of \prover on formal reasoning. We also conduct ablation studies and case studies to further validate the effectiveness of \dataset{}.
\subsection{Experiment settings}
\subsubsection{Model and Training}\label{sec:model_training}
Our training set mainly consists of three parts: the \dataset{}, Lean’s Mathlib (via the LeanDojo dataset), which is the common practive in training formal reasoning models in Lean, and other private synthetic theorem, which mainly comes from the autoformoalization effort of Lean Workbook. Several other models were trained with different training set settings for ablation. We follow the \textit{proofstep} objective used by \textit{GPT-f}~\cite{polu2020generative}, which generating a \texttt{PROOFSTEP} (a Lean tactic) given a \texttt{GOAL} (current Lean tactic state) and the current \texttt{DECLARATION} (the Lean theorem name to be proved): \texttt{DECL <DECLARATION> \textbackslash{n}GOAL <GOAL> \textbackslash{n}PROOFSTEP <PROOFSTEP>\textbackslash{n}}, as depicted in Fig.~\ref{dataexample}.
\begin{figure}[t]
  \centering
\begin{lstlisting}[language=LEAN,frame=single]
### Input
DECL MyNat.mul_pow
GOAL a b n : ℕ
⊢ (a * b) ^ n = a ^ n * b ^ n
### Output
PROOFSTEP induction n with t Ht
\end{lstlisting}
    \vspace{-0.4em}
  \caption{Examples of (input, output) pairs of our training prompt.}
  \label{dataexample}
\end{figure}

We used a global batch size of 512 and a learning rate of $10^{-5}$. We fine-tuned for 2 epoch to obtain the SFT model. For the learning rate, we used a warm-up in the first 3\% steps, followed by a cosine schedule decaying to zero. The whole fine-tuning process took around 6 hours on 32 A100 GPUs.
\subsubsection{Evaluation settings}

We utilized a standard methodology that iteratively performs a best-first search to generate tactics and validate intermediate proof steps within a formal proof until the proof is either finalized or resources are exhausted. During each generation step, a state $S_i$ is expanded by generating $S$ tactic candidates for it, with a maximum of $K$ expansions allowed in a single iteration. In this context, we select $S=32$ and $K=100$.

\textbf{State De-duplication}
\begin{wrapfigure}{R}{0.5\textwidth}
  \begin{center}
    \includegraphics[width=0.48\textwidth]{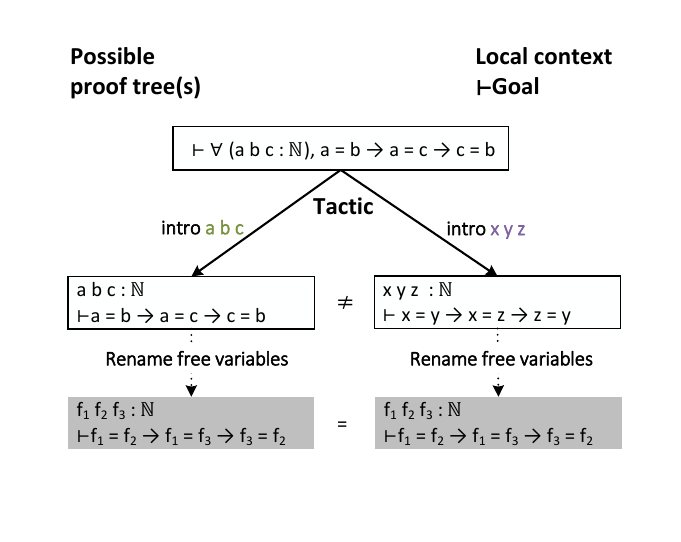}
  \end{center}
  \caption{De-duplication of intermediate proof states in tree search. Different tactics may result in states that are essentially the same. Renaming free variables and hypotheses based on their internal storage order in Lean's kernel provides a unified representation, helping identifying these states.}
  \label{fig:dedup}
\end{wrapfigure}
One of the most prevalent issues in the best-first search process is the highly duplicated states. 
This arises from the foundational nature of the Lean language, which is rooted in dependent type theory. In Lean, every proposition can be viewed as a valid type, and demonstrating that proposition is akin to discovering or constructing an element of that type, essentially proving that the proposition is \textit{inhabited}. Moreover, Lean considers two proofs of a proposition to be definitionally equivalent. Consequently, there is no inherent mechanism to distinguish if two (potentially partial) proofs lead to the same intermediate states. Given that numerous Lean tactics, such as \texttt{intro} and \texttt{have}, involve introducing new hypotheses with customized names, it is common for multiple intermediate states with essentially identical hypotheses but distinct names to emerge during the search process, as depicted in Fig.~\ref{fig:dedup}. This issue becomes more pronounced in extensive proof searches, where we have observed that over 50\% of intermediate states are duplicates. Failure to address this issue could result in significant computational inefficiencies.


To minimize computational wastage on duplicated intermediate states and improve the efficiency of the proof search process, we exploited Lean's runtime meta-programming facilities to provide additional information for de-duplication. We modified Lean so that for each intermediate state (internally stored as meta-variables and local declarations), their hypotheses are renamed based on their internal storage order. So that states with same hypotheses and goals could be identified, as shown in Fig.~\ref{fig:dedup}. 

\subsection{Main Results}
After testing with Lean 4 benchmarks aiming at different levels (high-school or undergraduate), difficulties (ordinary exercises or Olympiad problems), and taste (problem-solving vs. constructing a math system like in textbooks), we conclude that \prover{}exhibits versatile formal reasoning abilities compared to prior works by achieving state-of-the-art performance, proving the effectiveness of the diversity of \dataset{}.

\textbf{Results on miniF2F.} We first test the Lean 4 formal reasoning ability on miniF2F-Test
 and Validation dataset. MiniF2F~\cite{zheng2021minif2f} is a standard testing dataset for evaluating the performance of formal provers, containing 244 validation and 244 test problems, all stated in Lean. The range of problems varies from high-school competition questions to
 undergraduate-level theorem proofs, e.g. problems sampled from the
 MATH dataset~\cite{hendrycks2021measuringmathematicalproblemsolvingMATH}, high-school mathematical competitions (including AMC, AIME, and IMO) and some other manually crafted problems at the same difficulty. We use the version of miniF2F in Lean 4 released by the LeanDojo~\cite{yang2024leandojo} project, with adaptations to Lean 4.7.0.
 
We present the main experimental results in Tab.~\ref{tab:main_results}. From the table, we can observe that \prover achieves a cumulative accuracy rate of 63.9\% on miniF2F-Valid and 54.5\% on miniF2F-Test, suppressing all the baselines,
including DeepSeek-Prover~\cite{xin2024deepseekproveradvancingtheoremproving} which scores 60.2\% and 52.0\%, respectively. Specifically, \prover significantly outperforms all prior tree-search methods, including Hypertree Proof Search, which achieves only 58.6\% on miniF2F-valid and 41.0\% on miniF2F-test, demonstrating that the potential of tree search methods still remains to be fully explored and such method is comparable to generating whole proofs.

\textbf{Results on ProofNet.} ProofNet~\cite{azerbayev2023proofnetautoformalizingformallyproving} is a benchmark for testing the formal reasoning ability on undergraduate-level mathematics. It comprises 371 formal problems which are sourced from popular undergraduate pure mathematics textbooks and cover topics such as real and complex analysis, linear algebra, abstract algebra, and topology. We list results in Table~\ref{tab:main_results2}. Outperforming the benchmark's prior leader, ReProver \citep{yang2024leandojo}, which had a Pass@1 rate of 13.8\%, our model, \prover, achieves a significantly higher Pass@1 rate of 18.1\%(Fig.~\ref{tab:main_results2}). We have also discovered 24 proofs in ProofNet that currently do not have Lean proofs.  

\textbf{Results on PutnamBench.} PutnamBench~\cite{tsoukalas2024putnambenchevaluatingneuraltheoremprovers} is a benchmark comprising 640 theorems sourced from the Putnam Mathematical Competition, a renowned math competition in North America. These theorems are formalized in Lean 4, Isabelle, and partially in Coq. The benchmark is designed to test models' formal reasoning abilities in solving problems at a premier undergraduate mathematics level and is meticulously curated to prevent test-set leakage. We list results in Table~\ref{tab:main_results2}. Restricted to Lean 4 formalization, GPT-4, and COPRA \citep{thakur2023language} each solved one of the 640 problems, while ReProver failed to solve any. To our knowledge, the most effective method has been DSP \citep{jiang2022draft}, which operates in Isabelle and solved 4 problems with pass@10. As shown in Fig.~\ref{tab:main_results2}, \prover outperformed these results by solving 5 problems in a single pass without informal sketches and identified a solution for Putnam 1988 B2, a problem not yet reported to be solved by any ITP.
The generated proof is included in Tab.~\ref{tab:putnam_case}.
 
\begin{table}[th]
\begin{center}
\caption{
Comparing with state-of-the-arts on the miniF2F dataset.
}
\label{tab:main_results} 
\small
\begin{tabular}{lcccc}
\toprule
    Method & Model size & Pass & miniF2F-valid & miniF2F-test \\
     \midrule
    \multicolumn{5}{l}{\textit{Whole-Proof Generation Methods}} \\
    \midrule
    GPT-4-turbo 0409\citep{achiam2023gpt} & - & 64 & $25.4\%$ & $23.0\%$ \\
    \multirow{6}{*}{DeepSeek-Prover \citep{xin2024deepseekproveradvancingtheoremproving}} & \multirow{6}{*}{7B}  
     & $1$ & - & $30.0\%$ \\
     &  & $64$ & - & $46.3\%$ \\
     &  & $128$ & - & $46.3\%$ \\
     &  & $8192$ & - & $48.8\%$ \\
     &  & $65536$ & - & $50.0\%$ \\
     & & cumulative & \textit{60.2\%} & \textit{52.0\%} \\
    TheoremLlama \citep{wang2024theoremllamatransforminggeneralpurposellms} & - & cumulative & $36.5\%$ & $33.6\%$ \\
    \midrule
    \multicolumn{5}{l}{\textit{Tree Search Methods}} \\
    \midrule
     COPRA  (GPT-3.5) \citep{thakur2023language} & - & $1$ & - & $9.0\%$ \\
     COPRA (GPT-4) \citep{thakur2023language}& - & $1$ & - & $26.6\%$ \\
         DSP(Isabelle) \citep{jiang2022draft}   & 540B & $100$ & $42.6\%$ & $38.9\%$ \\
    \multirow{2}{*}{Proof Artifact Co-Training   \citep{han2021proof}} & \multirow{2}{*}{837M} & $1$ & $23.9\%$ & $24.6\%$ \\
    &  & $8$ & $29.3\%$ & $29.2\%$ \\
     ReProver   \citep{yang2024leandojo} & 229M & $1$ & - & $25.0\%$ \\
     Llemma   \citep{azerbayev2023llemma} & 7B & $1$ & $26.2\%$ & $26.2\%$ \\
     Llemma   \citep{azerbayev2023llemma} & 34B & $1$ & $27.9\%$ & $25.8\%$ \\

    \multirow{3}{*}{Curriculum Learning   \citep{polu2022formal}} & \multirow{3}{*}{837M} & $1$ & $33.6\%$ & $29.6\%$ \\
     &  & $8$ & $41.2\%$ & $34.5\%$ \\
     &  & $64$ & $47.3\%$ & $36.6\%$ \\
     \multirow{2}{*}{Hypertree Proof Search   \citep{lample2022hypertree}} & \multirow{2}{*}{600M} & cumulative & $58.6\%$ & - \\
     &  & $64$ & - & $41.0\%$ \\
     Lean-STaR ~\citep{lin2024leanstarlearninginterleavethinking}& 7B & 64 & - & $46.3\%$ \\
     InternLM2-Math~\cite{ying2024internlmmathopenmathlarge} & 7B & 1 & $29.9\%$ & $30.3\%$ \\
     InternLM2-Math-Plus~\cite{ying2024internlmmathopenmathlarge} & 7B & 1 &- & $43.4\%$  \\
     \midrule
     \multirow{2}{*}{\prover   } & \multirow{2}{*}{7B} &  $1$ & $59.8\%$ & $48.8\%$  \\
     && $64$ & $\textbf{63.9\%}$ & $\textbf{54.5\%}$\\    
    \bottomrule
\end{tabular}
\end{center}

\end{table}

\begin{table}[th]
\begin{center}
\caption{
Comparing with state-of-the-arts on the ProofNet~\citep{azerbayev2023proofnetautoformalizingformallyproving} and Putnam~\cite{tsoukalas2024putnambenchevaluatingneuraltheoremprovers} dataset.
}
\label{tab:main_results2} 
\small
\begin{tabular}{lccc}
\toprule
    Method & Model size & Pass & result \\
     \midrule
    \multicolumn{4}{l}{\textit{ProofNet~\citep{azerbayev2023proofnetautoformalizingformallyproving} benchmark}} \\
    \midrule
    ReProver   \citep{yang2024leandojo} & 229M & $1$ & 13.8\% \\
    \prover{} & 7B & 1 &\textbf{18.1\%}  \\
    \midrule
    \multicolumn{4}{l}{\textit{Putnam~\cite{tsoukalas2024putnambenchevaluatingneuraltheoremprovers} benchmark}} \\
    \midrule
    GPT-4 \citep{achiam2023gpt} & - & $10$  & 1/640\\
    COPRA (GPT-4) \citep{thakur2023language}& - & $10$ & 1/640 \\
    DSP(Isabelle) \citep{jiang2022draft}   & 540B & $10$ & 4/640 \\
    ReProver   \citep{yang2024leandojo} & 229M & $1$  & 0/640 \\
        \prover{} & 7B & 1 &\textbf{5/640}  \\  
    \bottomrule
\end{tabular}
\end{center}

\end{table}

\subsection{Ablation Studies}
\textbf{Data source ablation}
As described in Sec.~\ref{sec:model_training}, \prover is trained on \dataset{}, along with synthetic data including rule-based generated equations and inequalities and Lean-Workbook \citep{ying2024leanworkbooklargescalelean}, and human-written data extracted from Mathlib. To demonstrate the effectiveness of the \dataset{} dataset, we conducted a comparative analysis among various combinations of training data, as shown in Tab.~\ref{tab:ablation_data_source_miniF2F} and~\ref{tab:ablation_data_source_proofnet}. The results indicate that models trained with data extracted from GitHub significantly outperform those trained solely with Mathlib data and/or synthetic data. 
\begin{table}[th]
    \begin{center}
    \caption{
    Improvement in pass rates for miniF2F at pass@1 in models trained on formal proofs, with different data sources.
    }
    \label{tab:ablation_data_source_miniF2F} 
    \small
    \begin{tabular}{lcccc}
    
    \toprule
        Model & \#Tokens & miniF2F-valid & miniF2F-test \\
        \midrule
        Mathlib                                 &$0.131$B   & $44.3\%$ & $37.3\%$  \\
        Mathlib + \dataset{}                        & $0.269$B  & $44.3\%$ & $41.0\%$  \\
        Mathlib + synthetic theorems            & $1.286$B  & $58.2\%$ & $46.7\%$  \\
        Mathlib + \dataset{} + synthetic theorems & $1.424$B  & $59.8\%$ & $48.8\%$  \\
        \bottomrule
    \end{tabular}
    \end{center}
\end{table}

\begin{figure}[t]
	\centering
	\begin{minipage}[t]{0.48\textwidth}
		\centering
        \includegraphics[width=1.1\linewidth]{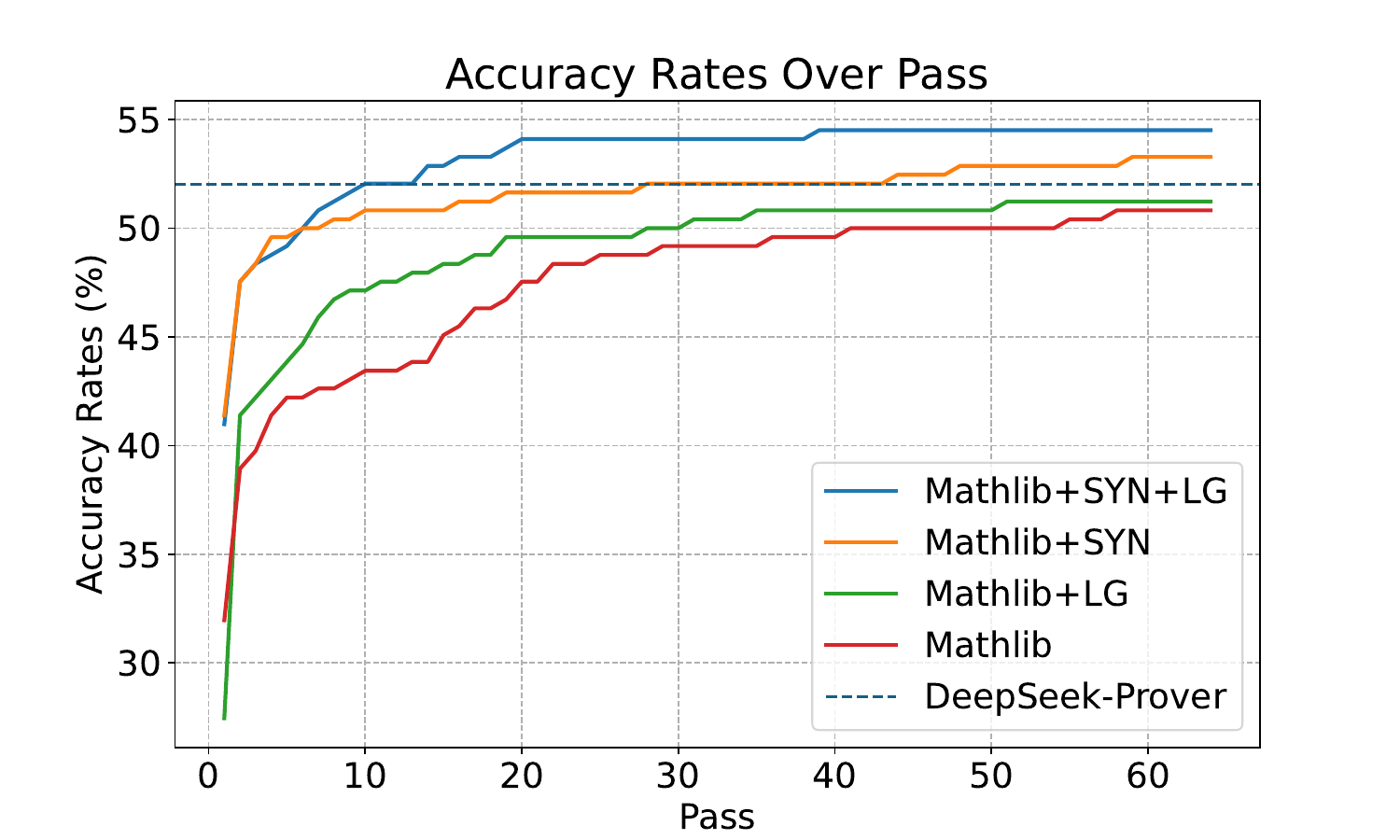}
		\caption{Improvement in pass rates for miniF2F-test at pass@64 in models trained with different data sources, where LG stands for \dataset{} and SYN stands for synthetic data.}
		\label{fig:test-pass64}
	\end{minipage}\hfill
	\begin{minipage}[t]{0.48\textwidth}
		\centering
		\includegraphics[width=1.1\linewidth]{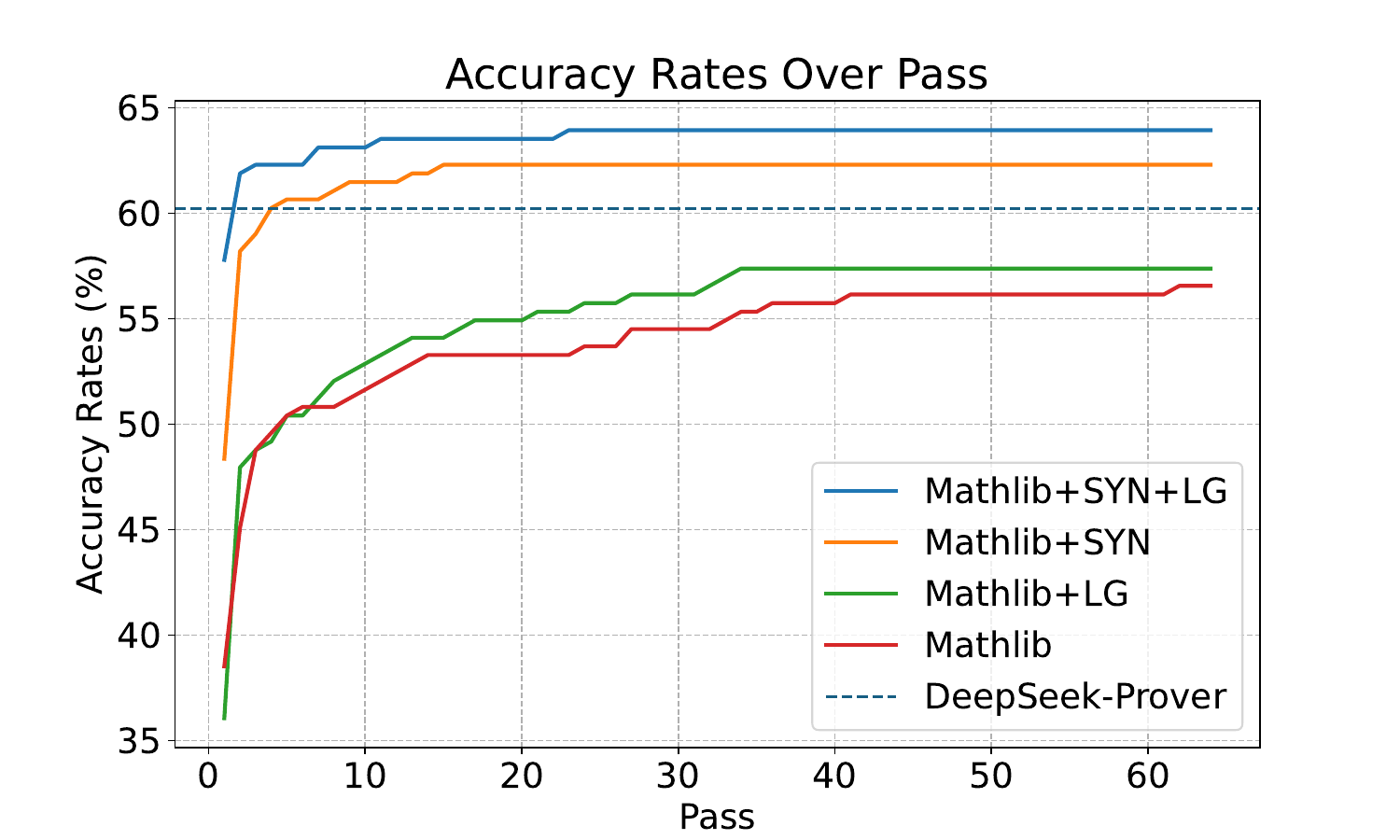}
		\caption{Improvement in pass rates for miniF2F-valid at pass@64 in models trained with different data sources.}
		\label{fig:valid-pass64}
	\end{minipage}
	\vspace{-0.15in}
\end{figure}

\begin{table}[thb]
    \begin{center}
    \caption{
    Improvement in pass rates for ProofNet at pass@1 in models trained on formal proofs, with different data sources.
    }
    \label{tab:ablation_data_source_proofnet} 
    \small
    \begin{tabular}{lcccc}
    
    \toprule
        Model & \#Tokens & ProofNet \\
        \midrule
        Mathlib                                 &$0.131$B   & $15.1\%$  \\
        Mathlib + \dataset{}                      & $0.269$B  & $16.2\%$  \\
        Mathlib + synthetic theorems            & $1.286$B  & $17.0\%$  \\
        Mathlib + \dataset{} + synthetic theorems & $1.424$B  & $18.1\%$  \\
        \bottomrule
    \end{tabular}
    \end{center}
\end{table}
\textbf{The effectiveness of multiple inferences.} We then focused on how \dataset{} affects formal reasoning performance when scaling evaluation (i.e. with more pass time). Since the formalization system will tell us whether a problem is solved, there is no reason to restrict the model to a single pass when pursuing maximum performance. By extending the evaluation budget until the performance improvement is marginal, as shown in Fig.~\ref{fig:test-pass64} and~\ref{fig:valid-pass64}, we observe that \dataset{} improves the models' maximum performance.

We proceeded with the evaluation of each model using temperatures 0.7 and 1.0, each with 32 independent inferences, instead of beam search which we had chosen for the Pass@1 evaluation. Therefore, the initial accuracy rates of the first round may be lower than in the Pass@1 evaluation. \prover achieved an accumulated pass rate of 54.5\% for miniF2F-test and 63.9\% for miniF2F-valid, surpassing the baseline trained without \dataset{}, which had pass rates of 53.2\% and 62.3\%, respectively. The same trend was observed for models trained solely on Mathlib and on Mathlib+\dataset{}. We have also discovered a proof for IMO 1983 P6, which, to our best knowledge, has not been proved in Lean 4 before (also not be solved in compfiles\footnote{\url{https://github.com/dwrensha/compfiles/blob/main/Compfiles/Imo1983P6.lean}}). Examples of proved theorems can be found in Sec.~\ref{sec:case}.

\section{Conclusion}

In this paper, we introduce \dataset{}—a dataset comprising a large-scale collection of formal data extracted from open Lean 4 repositories on GitHub, which includes 28,597 theorems and 218,866 tactics. We then train \prover using this dataset, which is the state-of-the-art model performance on Lean 4 formal reasoning.
We also train various models with \dataset{} to evaluate the formal reasoning performance that can be achieved by training on our dataset. Notably, we find that models trained on \dataset{} exhibit performance improvements in formal reasoning across various fields and difficulty levels, demonstrating that a well-extracted, diverse dataset can enhance model performance on a range of reasoning tasks. We hope that by opening \dataset{}, we can assist the community in better exploiting the under-utilized information in raw corpora and in improving mathematical reasoning capabilities.
\nocite{wei2022finetunedlanguagemodelszeroshot}
\bibliography{refs}
\bibliographystyle{plain}




\newpage

\newpage
\appendix

\section{Case study}\label{sec:case}
This section presents case studies to demonstrate the performance of our methods.
\begin{tcolorbox}[
colback=white!10!white,
colframe=purple!75!purple,
title=Case 1: Binomial Coefficients
]\label{tab:minif2f_induction_cases}
\textcolor{blue}{Natural Language problem:} Show that for positive integers $n$ and $k$ with $k \leq n$, we have $\binom{n}{k} = \binom{n-1}{k} + \binom{n-1}{k-1}$. 
\\


\begin{lstlisting}[language=LEAN]
theorem numbertheory_nckeqnm1ckpnm1ckm1 (n k : ℕ) (h₀ : 0 < n ∧ 0 < k) (h₁ : k ≤ n) : Nat.choose n k = Nat.choose (n - 1) k + Nat.choose (n - 1) (k - 1) := 
by
    induction n 
    all_goals cases' k
    all_goals simp [choose, h₀.1.ne', tsub_eq_zero_of_le (Nat.succ_le_of_lt h₀.2), add_zero] at *
    rw [add_comm]
\end{lstlisting}

\end{tcolorbox}

In this case, \prover exhibits its ability to do simple inductions. In addition, it is capable to solving high-school level number theory problems as well as the algebra problems.
\begin{tcolorbox}[
colback=white!10!white,
colframe=purple!75!purple,
title=Case 2: Putnam 1988 B2
]\label{tab:putnam_case}
\textcolor{blue}{Natural Language problem:} Prove or disprove\footnote{We search for two directions (prove/disprove) and only present the correct case in our formalization.}: If $x$ and $y$ are real numbers with $y \geq 0$ and $y(y+1) \leq (x+1)^2$, then $y(y-1) \leq x^2$. 
\\


\begin{lstlisting}[language=LEAN]
theorem putnam_1988_b2: 
(∀ x y : ℝ, (y ≥ 0 ∧ y * (y + 1) ≤ (x + 1) ^ 2) → (y * (y - 1) ≤ x ^ 2)) ↔ True :=
by
    refine ⟨fun _ ↦ trivial, fun _ x y hy ↦?_⟩
    ring_nf at hy ⊢
    nlinarith [sq_nonneg (x - y)]
\end{lstlisting}

\end{tcolorbox}

The generated proof first breaks up the conjunction into two implications, each becoming a subgoal. The latter goal is trivial and automatically closed. \prover  then uses automated tactics such as \texttt{ring\_nf} and \texttt{nlinarith} to close the other subgoal. The key step in this proof is to provide appropriate hints (i.e. \texttt{sq\_nonneq}) to the underlying automated tactics. 

\begin{tcolorbox}[
colback=white!10!white,
colframe=purple!75!purple,
title=Case 3: IMO 1964 P1(2)
]
\textcolor{blue}{Natural Language problem:} Prove that there is no positive integer $n$ for which $2^n+1$ is divisible by $7$.
\\


\begin{lstlisting}[language=LEAN]
theorem imo_1964_p1_2 (n : ℕ) : ¬7 | 2 ^ n + 1 := 
by
    intro h
    rw [← Nat.mod_add_div n 3] at h
    rw [Nat.dvd_iff_mod_eq_zero] at h
    have h₁ : n % 3 < 3 := Nat.mod_lt n three_pos
    interval_cases n % 3 
    all_goals simp [pow_add, pow_mul, Nat.add_mod, Nat.pow_mod, Nat.mul_mod] at h ⊢
\end{lstlisting}
\label{tab:imo_case}
\end{tcolorbox}
IMO 1964 P1 is a composite problem, consisting of a relatively simple first part and a more complex second part, with the conclusion of the first part being used in the second. In the formalized version, we omitted the lemma from the first part, which actually increases the difficulty of the problem. The key to solving the problem lies in finding the cycle of $2^n \mod 7$, which is (1,2,4). The \textit{prover} identified that the length of this cycle is 3 and conducted a case analysis based on 3 (having previously proved some necessary premises), successfully solving the problem.

\begin{tcolorbox}[
colback=white!10!white,
colframe=purple!75!purple,
title=Case 4  : Pough 3.2.8
]\label{tab:proofnet_case}
\textcolor{blue}{Natural Language problem:} Prove that if $H$ and $K$ are finite subgroups of $G$ whose orders are relatively prime then $H \cap K=1$. 
\\


\begin{lstlisting}[language=LEAN]
theorem exercise_Dummit_3_2_8 {G : Type*} [Group G] (H K : Subgroup G)
  [Fintype H] [Fintype K]
  (hHK : Nat.Coprime (card H) (card K)) :
  H ⊓ K = ⊥ := 
by
    rw [eq_bot_iff_forall]
    rintro x ⟨hx : x ∈ H, hx' : x ∈ K⟩
    have : x ∈ H ⊓ K := ⟨hx, hx'⟩
    rw [inf_eq_bot_of_coprime hHK] at this
    exact Subgroup.mem_bot.mp this
\end{lstlisting}

\end{tcolorbox}

\prover is also capable of solving undergraduate problems that needs to reason over a larger repository of premises, in this case the knowledge of group and co-prime. The case exhibits \dataset{}'s effectiveness on versatile mathematic reasoning tasks.

\end{document}